\documentclass[letterpaper, 10 pt, conference]{ieeeconf}  

\usepackage{amsmath}
\usepackage{algorithm}
\usepackage{algpseudocode}
\usepackage{amsfonts}
\usepackage{graphicx}
\usepackage{cite} 
\usepackage{caption}
\usepackage[scr=esstix]{mathalpha}
\usepackage{placeins}

\newcommand{\vect}[1]{\mathbf{#1}} 

\newcommand{\bK}{{\bf K}}
\newcommand{\bS}{{\bf S}}

\newcommand{\bQ}{{\bf Q}}

\newcommand{\bxi}{\boldsymbol{\xi}}

\newcommand{\bzeta}{\boldsymbol{\zeta}}

\newcommand{\bx}{\mathbf{x}}
\newcommand{\br}{\mathbf{r}}

\newcommand{\bv}{\mathbf{v}}
\newcommand{\bu}{\mathbf{u}}

\newcommand{\bez}{{\bf \hat{e}}_{z}}
\newcommand{\ben}{{\bf \hat{e}}_{n}}
\newcommand{\bet}{{\bf \hat{e}}_{t}}

\IEEEoverridecommandlockouts                              

\title{\LARGE \bf Real-Time Maneuver Planning for Fixed-Wing UAVs in Unsteady Flows Using a GPU-Accelerated Vortex Particle Model} 

\author{Ashwin Gupta$^{1}$, Kevin Wolfe$^{2}$, Gino Perrotta$^{2}$, and Joseph Moore$^{1,2}$
\thanks{
$^{1}$Johns Hopkins University Whiting School of Engineering \newline \hspace*{1.6em} {\tt\small \{agupt139,jlmoore@\}@jh.edu}
\newline \hspace*{0.8em}
$^{2}$Johns Hopkins University Applied Physics Lab \newline \hspace*{1.6em}
{\tt\small \{Kevin.Wolfe,Gino.Perrotta\}@jhuapl.edu} 
}
}

\renewcommand{\baselinestretch}{0.98}

\begin{document}

\maketitle
\thispagestyle{empty}
\pagestyle{empty}

\begin{abstract}

Unsteady aerodynamic effects can have a profound impact on aerial vehicle flight performance, especially during agile maneuvers and in complex aerodynamic environments. In this paper, we present a real-time planning and control approach capable of reasoning about unsteady aerodynamics. Our approach relies on a lightweight GPU-accelerated vortex particle model (VPM) and a sampling-based policy optimization strategy capable of leveraging the VPM for predictive reasoning. Through hardware experiments, we show that by replanning with our unsteady aerodynamics model, we can improve the performance of a post-stall fixed-wing perching maneuver in the presence of unsteady environmental flow disturbances. 
\end{abstract}

\section{INTRODUCTION}

For more than two decades, the research community has sought to improve the flight control performance of uncrewed aerial vehicles (UAVs). Early research explored control strategies for autonomous helicopter control using nonlinear model predictive control (NMPC)\cite{kim2002nonlinear} and later on more aggressive maneuvering with multi-rotor UAVs \cite{mellinger2011minimum}. More recently, researchers have explored control strategies for agile fixed-wing control \cite{bulka2018autonomous,basescu2020icra}, as well as hybrid VTOL aircraft \cite{saeed2015review}. While researchers have investigated many different types of control strategies, from model-based control \cite{nguyen2021model} to reinforcement learning \cite{fagundes2024machine}, most approaches have leveraged quasi-static, lumped-parameter aerodynamics models. 

However, as the performance envelopes of UAVs continue to increase, unsteady aerodynamic effects will start to play a more impactful role. This is particularly true for applications such as urban air mobility, where vortex shedding off of nearby structures leads to an unsteady environmental flow field \cite{sutherland2016urban}, or landing in the turbulent wake of a ship \cite{misra2019modeling}.

In this paper, we propose a control algorithm that can reason about unsteady aerodynamic effects produced by the vehicle itself as well as the environment. Our approach constructs a lightweight computational fluid dynamics (CFD) model, known as a vortex particle model (VPM), to explicitly model the unsteady flow field. To enable real-time planning, we parallelize both the VPM and the dynamics rollouts. To address the challenges with model differentiability, we leverage a stochastic trajectory optimization approach, model predictive path integral control (MPPI), and build a local trajectory tracking controller directly from samples.

Through hardware experiments, we evaluate the performance of our approach on a two-dimensional post-stall perching maneuver \cite{moore2014robust} subject to a vortex disturbance. Because our vortex particle model represents the fluid state using point vortices, a planarized ring vortex disturbance can be naturally added to the fluid state. We use a large vortex cannon to generate this ring vortex disturbance and detect the vortex using offboard pressure sensors local to the perch site. Our results show improved perching performance in the presence of an unsteady environmental disturbance.

Our primary contributions are:
\begin{itemize}
\item A parallelized vortex particle model suitable for real-time planning.
\item A framework for real-time planning and control with a computationally-tractable VPM.
\item Hardware experiments demonstrating improved maneuver performance for a perching UAV in unsteady aerodynamic environments.
\end{itemize}
While our approach is currently restricted to two-dimensional maneuvers with offboard sensing and compute, we view it as a meaningful step towards CFD-in-the-loop control.

\begin{figure}[!t]%
  \centering
    \includegraphics[trim={1cm 0 0 2cm},clip, width=1\columnwidth]{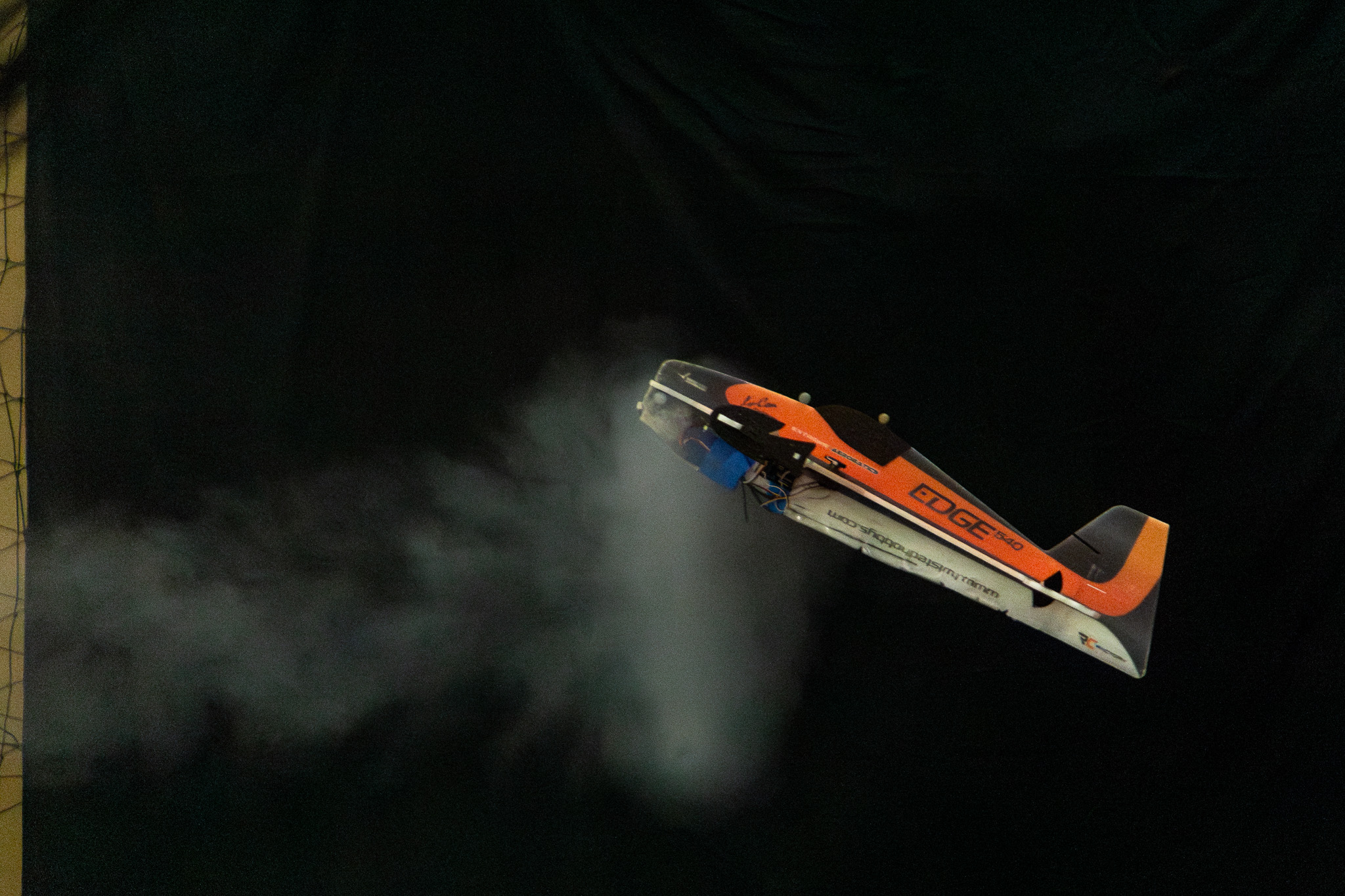}
    \caption{Still image of the glider-vortex interaction}
  \label{fig:cover}
\end{figure}

\section{RELATED WORK}
Model-based controllers have achieved impressive performance for both fixed-wing and multi-rotor UAVs relying primarily on quasi-steady, lumped-parameter models \cite{moore2014robust,mahony2012multirotor}. Fixed-wing UAV controllers have demonstrated aggressive, post-stall maneuvering \cite{basescu2020icra} while relying on quasi-steady flat plate models, while quadcopter control can achieve acrobatic maneuvers using simple, differentially flat models \cite{mellinger2011minimum}. In \cite{khan2016modeling}, researchers leverage the unsteady lumped parameter high angle-of-attack model proposed in \cite{goman1994state} to create a higher fidelity fixed-wing UAV model. 

\subsection{Linear Control with CFD}
To further increase the fidelity of the models used for aircraft control, researchers have explored leveraging computational fluid dynamics (CFD). Because CFD is not computationally efficient enough to run online, most approaches have focused on fitting model parameters to data generated offline. In particular, researchers have sought to fit dynamic derivatives associated with unsteady aerodynamic effects \cite{ronch2012evaluation,gortz2007towards}. While most approaches have explored grid-based CFD approaches, recently \cite{gino-dvs} explored using a VPM. While more computationally efficient than grid-based CFD, this approach also relied on an offline generation of a local time-varying linear policy.

\subsection{Model-Predictive Control with Unsteady Models}
In recent years, model predictive control has proven to be an effective approach for controlling UAVs under challenging conditions (e.g., \cite{basescu2020icra,mathisen2021precision}). In some cases, researchers have sought to incorporate unsteady flow models into an MPC framework. \cite{misra2018stochastic} and \cite{misra2019output} modeled turbulent flows and ship wakes with stochastic models and leveraged stochastic MPC to enable shipboard landings.  In \cite{ngo2016model}, researchers integrated a high-fidelity aerodynamics model and CFD-based wake model to achieve MPC-based landing with an autonomous rotorcraft. In \cite{lorenzetti2020uav}, the authors leveraged reduced-order modeling technique to incorporate a CFD-based model into MPC for glideslope tracking in a ship airwake. 

\subsection{Machine Learning Approaches}
Machine learning has also proven to be a useful tool to capture complex, unsteady aerodynamic effects. Researchers have used machine learning to augment aerodynamics models using real-world data, for both multi-rotor UAVs \cite{bauersfeld2021neurobem} and fixed-wing UAVs \cite{basescu2023precision}. Researchers have also used CFD to fit improved aerodynamics models. In \cite{fukami2023}, authors propose a novel learning framework for learning a latent space to learn higher fidelity models from data. \cite{Liu_2025} builds on this model to apply model-based reinforcement learning to control airfoil in a highly disturbed flow. Several approaches leveraged specialized flow sensing combined with reinforcement learning to enable control in unsteady environmental flows in hardware. \cite{simon2023flowdrone} introduced a fast-response hot-wire anemometer to improve quadcopter control performance via reinforcement learning. \cite{roberts2012control} and \cite{bauersfeld2025low} have explored the use of full particle image velocimetry (PIV) with reinforcement learning to improve performance in unsteady flow fields.  


In this paper, we present an approach for real-time planning and control that leverages a computationally lightweight VPM ``in-the-loop'' to improve performance in highly disturbed flows. To our knowledge, this is the first demonstration of using a CFD model in-the-loop for real-time control of a UAV.


\section{Unsteady Aerodynamics Model} \label{sec:unsteady}
To model the unsteady fluid dynamics associated with a fixed-wing UAV, we combine a VPM with the rigid body dynamics of a planarized glider. In particular, we use the VPM to model the unsteady aerodynamic effects of the wing, while using a lumped parameter model to capture the forces applied by the elevator.

A number of computational methods exist for modeling the unsteady aerodynamic forces acting on a wing. Many grid-based, Eulerian CFD approaches can model the unsteady fluid effects with considerable accuracy. However, these high-fidelity CFD approaches are associated with a considerable computational cost, even when using high-performance computing and are not suitable for inclusion in a real-time control loop \cite{smith2025advances}. Other machine learning-based approaches or reduced-order modeling approaches based on these Eulerian flow descriptions present challenges with regard to generalization in the presence of out-of-distribution maneuvers or environments. For this reason, we employ a VPM based on \cite{Katz_Plotkin_2001} and explore using a Lagrangian parameterization of the flow. By representing the flow state as a set of discrete position particles that convect with the flow and are characterized by a scalar vorticity value, we can achieve a computationally tractable model of the unsteady aerodynamics.

\subsection{Vortex Particle Model}
The VPM assumes inviscid and incompressible flow. Since the flow is incompressible, the velocity field ${\bf v}$ has the property $\nabla \cdot \bv=0$. If we define vorticity as $\bzeta = \nabla \times \bv$, then in two dimensions, the velocity field at position $\br\in\mathbb{R}^2$ can be written as 
\begin{align}
\bv(\br) = -\frac{1}{2\pi}\int_R \frac{(\br-\br')\times\bzeta(\br')}{|\br-\br'|^2}dA'
\label{eq:velfield}
\end{align}
where the vorticity is being integrated over some region R. For a set of $N_{\textit{v}}$ discrete point vortices $\textit{v}\in \mathcal{V}$ defined by position $\br_{\textit{v}}$  and circulation $\Gamma_{\textit{v}}$, Eq. \ref{eq:velfield} becomes 
\begin{align}
\bv(\br) = -\frac{1}{2\pi}\sum_{\textit{v}\in\mathcal{V}} \frac{(\br-\br_{\textit{v}})\times\Gamma_{\textit{v}} \bez}{|\br-\br_{\textit{v}}|^2},
\end{align}
where $\bez$ is the unit vector normal to the plane.
This can be more compactly written as 
\begin{align}
\bv(\br) = \sum_{\textit{v}\in\mathcal{V}} \bK_{\textit{v}}(\br) \Gamma_{\textit{v}},
\end{align}
where $\bK_{\textit{v}}(\br)$ is the vortex kernel given as 
\begin{align}
\bK_{\textit{v}}(\br)=-\frac{1}{2\pi r^{2}}
    \begin{bmatrix}
        0 & 1 \\
        -1 & 0
    \end{bmatrix}
    \Delta \vect{r}, 
    \label{eq:kernel}
\end{align}
for $\Delta \vect{r} = \br-\br_{\textit{v}}$ and $r = |\Delta \vect{r}|$. 

In some cases, these vortex kernels are modified to account for the finite core observed in a real viscous vortex. The model in \cite{leishman2006principles} scales the kernel in Eq. \ref{eq:kernel} as follows
\begin{align} 
\bK_{\textit{v}}'(\br)=
      \frac{\left( \frac{r}{r_{\mathrm{core}}} \right)^{2}}{\sqrt{1 + \left( \frac{r}{r_{\mathrm{core}}} \right)^{4}}}\bK_{\textit{v}}(\br), 
    \label{eq:regularized_kernel}
\end{align}
where $r_{core}$ is a prespecified constant core radius. 

We use vortex particles to represent both the vorticity in the wing wake \emph{and} the vorticity bound in the wing itself. In particular, we use Eq. \ref{eq:kernel} to model the bound vortices, and Eq. \ref{eq:regularized_kernel} to model wake vortices. To model the wing, we discretize the surface into $N_c$ equally-spaced collocation points (including the leading and trailing edge) and place $N_b = N_c-1$ bound vortices between these collocation points. At each of the collocation points, we also enforce the ``no through flow'' condition, 
\begin{align}
\bv_{c_i} \cdot \ben = 0,
\label{eq:no-through-flow}
\end{align}
where $\bv_{c_i}$ is the flow velocity at collocation point $i$ and $\vect{\hat{e}}_n$ is the unit vector normal to the surface. 

To model vortex shedding, we add one new leading-edge vortex (LEV) and trailing-edge vortex (TEV) to the wake at every simulation time step with position $\br_{\text{LEV}} $ and $\br_{\text{TEV}}$; and circulation $\Gamma_{LEV}$ and $\Gamma_{TEV}$, respectively. Following \cite{Katz_Plotkin_2001}, $\br_{\text{LEV}}$ and $\br_{\text{TEV}}$ are chosen to be $\pm \delta_s \bet$ from the leading and trailing edge collocation point respectively, where $\bet$ is the unit vector tangent to the surface and $\delta_s$ is 2\% of the chord. 

The unknown circulations of the $N_b$ bound and two shed vortices lead to $N_b+2$ unknown variables for the $N_b+1$ no-through-flow constraints at each collocation point. To close this system of equations, we use Kelvin's circulation theorem
\begin{align}
\frac{d}{dt}\sum_{\textit{v}\in \mathcal{V}} \Gamma_\textit{v} = 0,
\label{eq:kelvin}
\end{align}
which states that the total circulation is conserved. In our case, the flow is initialized with zero circulation which is maintained for all time. Here, $\mathcal{V}$ is the set of all vortices.

Let the set of bound vortices be $\mathcal{B}=\{b_1,b_2\dots b_{N_b}\}$, the set of wake vortices be $\mathcal{W}=\{\textit{w}_1,\textit{w}_2\dots \textit{w}_{N_\textit{w}}\}$, and the set of shed vortices be $\mathcal{E}= \{ \mathscr{e}_\text{TEV}, \mathscr{e}_\text{LEV}\}$, where the set of ``surface'' vortices is given as $\mathcal{S}=\mathcal{B}\cup \mathcal{E}$.
If $\br_{c_i}$ is the $i^{th}$ control point, and we let
\begin{align} \label{eq:system}
&\boldsymbol{\mathcal{K}}=\nonumber\\
&\begin{bmatrix} 
\vect{K}_{b_1}(\br_{c_1}) \cdot \vect{n} &\ldots  \vect{K}_{\mathscr{e}_\text{LEV}}'(\br_{c_1})\cdot \vect{n} & \ldots \vect{K}_{\mathscr{e}_\text{TEV}}'(\br_{c_1}) \cdot \vect{n} \\
\vect{K}_{b{_1}}(\br_{c_2}) \cdot \vect{n} &\ldots  \vect{K}'_{\mathscr{e}_\text{LEV}}(\br_{c_2}) \cdot \vect{n} & \ldots \vect{K}'_{\mathscr{e}_\text{TEV}}(\br_{c_2}) \cdot \vect{n}\\
\ldots & \ldots  & \ldots \\
1 & \ldots & 1  \end{bmatrix},
\end{align}
\begin{align}
{\bf \Gamma}_\mathcal {S} =
\begin{bmatrix}  \Gamma_{b_1} & \Gamma_{b_2} &...& \Gamma_{b_{N_b}}  &\Gamma_{\mathscr{e}_\text{LEV}} & \Gamma_{\mathscr{e}_\text{TEV}}
\end{bmatrix}^T,
\end{align}
and
\begin{align}
\vect{b} = 
\begin{bmatrix}  
-\bv^{\mathcal{W}}_{c_1} \cdot \vect{n} & \ldots & -\bv^{\mathcal{W}}_{c_{N_c}} \cdot \vect{n} & - \sum_{\textit{w}\in\mathcal{W}} \Gamma_\textit{w}
\end{bmatrix}^T,
\end{align}
we can use Eq. \ref{eq:no-through-flow} and Eq. \ref{eq:kelvin} to write
\begin{align}
{\bf \Gamma}_\mathcal {S} = \boldsymbol{\mathcal{K}}^{-1}{\bf b}.
\label{eq:circulation}
\end{align}
Here $\bv_{c_i}^{\mathcal{W}}$ is the velocity at the control points due to the wake vortices given as $\bv_{c_i}^{\mathcal{W}} = \tilde{\bv}(\br_{c_i})$ where 
\begin{align} \label{eq:wake-contrib}
\tilde{\bv}(\br_s) = \sum_{\textit{w}\in \mathcal{W}} \bK_\textit{w}^{'} (\br_{s})\Gamma_{\textit{w}}-\bv - \omega \bez \times (\br_g-\br_{s}).
\end{align}
$\bv$ is the translational velocity due surface motion, $\omega$ is the angular velocity of the surface, $\br_g$ is the position of the surface origin, and $\br_s$ is a point on the surface. 
At every time step, Eq. \ref{eq:circulation} provides a means of updating the bound circulation, as well as the circulation of the leading and trailing edge vortices. To compute the pressure on the surface, we use the unsteady Bernoulli equation \cite{Katz_Plotkin_2001}
\begin{align}
\Delta p_i = \rho \bigg [ \frac{\Gamma_{b_i}}{l_b}\tilde{\bv}(\br_{b_i})\cdot {\bf e}_t + \frac{\partial}{\partial t}\sum_{j=1}^i \Gamma_{b_j} + \frac{\partial}{\partial t}\Gamma_{\text{LEV}} \bigg],
\end{align}
where $l_b$ is the distance between bound vortices and $\br_{b_i}$ is the position of bound vortex $i$.

The force and moment on the wing then becomes 
\begin{align}
{\bf F}_w &= \sum_i^{N_b} l_b s\Delta p_i \ben,\label{eq:wingforce}\\
M_w &= \sum_i^{N_b} l_b s\Delta p_i (\br_{b_i}-\br_g)\cdot \bet\label{eq:wingmoment},
\end{align}
where $s$ is the wing span.

After the new leading and trailing edge vortex circulations are calculated, they are included in the wake so that $\mathcal{W} = \mathcal{W}\cup\mathcal{E}$, and dynamics of the wake vortices are modeled as
\begin{align} \label{eq:convect}
\dot{\br}_{\textit{w}_i} = \sum_{\textit{w}\in (\mathcal {W} - \textit{w}_i)}\bK_\textit{w}^{'} (\br_{\textit{w}_i}) \Gamma_\textit{w} +\sum_{\textit{b}\in \mathcal {B}}\bK_\textit{b} (\br_{\textit{w}_i}) \Gamma_b.
\end{align}
To model viscous dissipation, the wake vortex circulation dynamics are then given as 
\begin{align}
    \dot{\Gamma}_{\textit{w}_i} = -\gamma \Gamma_{\textit{w}_i},
\end{align}
where $0<\gamma<1$.

To prevent the accumulation of an excessive number of wake particles, we combine particles when the total number exceeds a predetermined threshold, $N_{\textit{w},\text{max}}$. The two oldest particles are merged into a single particle at their average position, and the strength of the new particle is equal to the sum of their strengths. This approach also preserves conservation of circulation.

\begin{figure}[tbh]%
  \centering
    \includegraphics[clip, trim={0 1cm 0 2.1cm}, width=1.0\columnwidth]{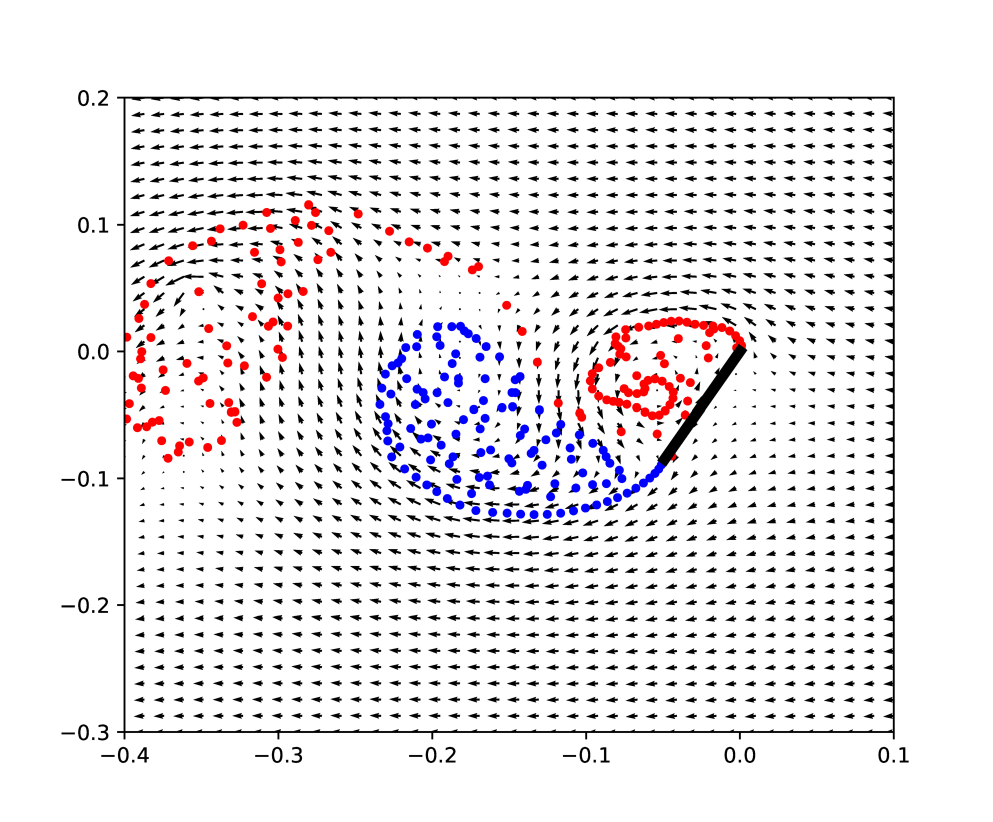}
    \caption{LEV formation and shedding with the VPM. Red and blue dots depict wake vortices with counter-clockwise and clockwise orientations respectively. The black line depicts the wing chord. Bound wing vortices are not pictured.}
  \label{fig:vpm}
\end{figure}

To model flow separation, we do not enable vortex shedding until after the glider exceeds a critical angle of attack. While more sophisticated approaches exist based on models of leading-edge suction \cite{LESP}, we choose this simpler method to improve overall computational efficiency. 
A visualization of the VPM at high angles of attack is shown in Figure \ref{fig:vpm}.

\subsection{Glider Dynamics}
We construct our nonlinear dynamics model based on the two dimensional flat plate glider model detailed in \cite{moore2014robust} consisting of two lifting surfaces: a wing and a single elevator control surface at the tail. The glider state is the position, pitch, elevator deflection, velocity, and pitch rate given as
$\vect{x}_g = \begin{bmatrix} x & z & \theta & \phi & \dot{x} & \dot{z} & \dot{\theta}\end{bmatrix}^T$. The elevator normal vector is given as ${\bf n}_e = \begin{bmatrix} -s_{\theta+\phi} & c_{\theta+\phi} \end{bmatrix}^T$, where $s_\gamma=\sin(\gamma)$ and $c_\gamma=\cos(\gamma)$. The positions and velocities of the wing and elevator are given as
\begin{align}
\bx_w &= \begin{bmatrix} x - l_w c_\theta \\ z -
  l_w s_\theta \end{bmatrix}, \quad \bx_e =
  \begin{bmatrix} x - l c_\theta - l_e c_{\theta+\phi} \\ z -
  l s_\theta - l_e s_{\theta+\phi} \end{bmatrix}, 
\\ \nonumber
\dot\bx_w &= \begin{bmatrix} \dot{x} + l_w \dot\theta s_\theta \\ \dot{z} -
  l_w \dot\theta c_\theta \end{bmatrix}, \quad \dot\bx_e =
  \begin{bmatrix} \dot{x} + l\dot\theta s_\theta + l_e
    (\dot\theta+\dot\phi) s_{\theta+\phi} \\ \dot{z} -
  l\dot\theta c_\theta - l_e(\dot\theta + \dot\phi)c_{\theta+\phi} \end{bmatrix}
\end{align}  where $l_e$ is the length of the elevator, $l$ is the body length, and $l_w$ is the length of the wing.
The elevator force is computed from flat plate theory as
\begin{align}
&\alpha_e =
\theta+\phi - \tan^{-1}(\dot z_e,\dot x_e) \\
&{\bf F}_e = \frac{1}{2}\rho |\dot{\bx}_e |^2 S_e C_n\bf{n}_e 
\end{align}
where $\alpha_e$ is the elevator angle-of-attack, $\rho$ is the density of air, and $S_e$ is the surface area of the tail control surface, and $C_n = 2\sin{\alpha_e}$. The wing force ${\bf F}_w$ and $M_w$ come from the VPM (i.e., Eq. \ref{eq:wingforce} and Eq. \ref{eq:wingmoment}). When combined with the VPM aerodynamics, the two-dimensional unsteady glider model becomes:
\begin{align} 
&m\begin{bmatrix} \ddot{x} \\ \ddot{z} \end{bmatrix} = {\bf F}_w+{\bf F}_e-\begin{bmatrix} 0\\ mg  \end{bmatrix} \\ 
&I\ddot\theta = \begin{bmatrix} l_w \\ 0\end{bmatrix} \times {\bf F}_w + \begin{bmatrix} -l-l_e c_{\theta} \\ -l+l_e s_{\theta} \end{bmatrix} \times {\bf F}_e + M_w.
\end{align}
Together with the states of VPM, the full state of the unsteady glider model is 
\begin{align}
\bx = 
\begin{bmatrix} \bx_g^T & \br_{{\textit{w}_1}}^T & \br_{{\textit{w}_2}}^T & \ldots & \br_{\textit{w}_{N_{\textit{w},\text{max}}}}^T &
& {\boldsymbol{\Gamma}_\mathcal{B}}^T & 
{\boldsymbol{\Gamma}_{\mathcal{W}}}^T
\end{bmatrix}^T
\end{align}
where 
${\bf \Gamma}_\mathcal {B} =
\begin{bmatrix}  \Gamma_{\textit{b}_1} & \Gamma_{\textit{b}_2} &...& \Gamma_{b_{N_{b}}} \end{bmatrix}^T$ and
${\bf \Gamma}_\mathcal {W} =
\begin{bmatrix}  \Gamma_{\textit{w}_1} & \Gamma_{\textit{w}_2} &...& \Gamma_{\textit{w}_{N_{\textit{w},\text{max}}}} \end{bmatrix}^T$. The state has a total size of $7+N_b + 3 \times N_{\textit{w},\text{max}}$. The control input is the elevator rate, where $\bu = \dot{\phi}$. 

\subsection{GPU Parallelization}
\begin{figure}[tbh]%
  \centering
    \includegraphics[width=1.0\columnwidth]{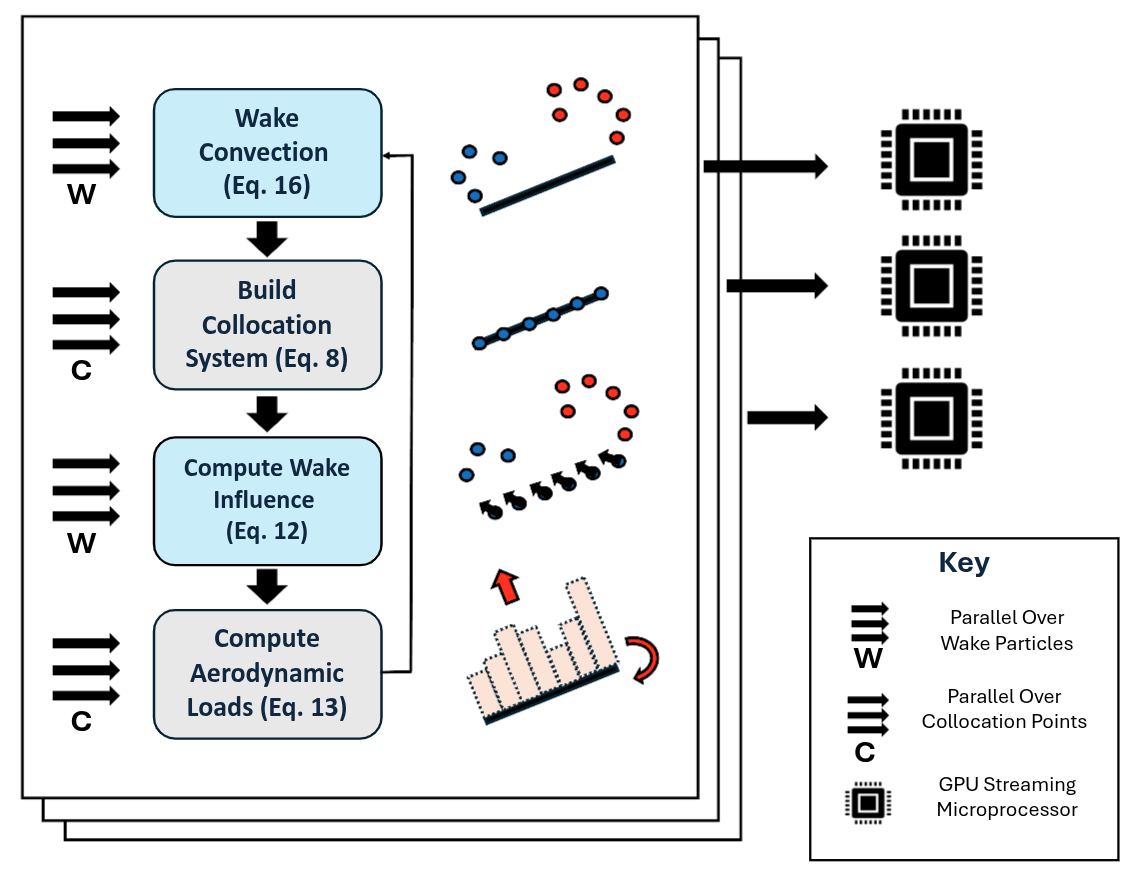}
    \caption{
   Each card represents a CUDA thread block, which contains 256 parallel working threads. Thread blocks are distributed by the CUDA runtime across dozens of streaming multiprocessors on the GPU. Each VPM instance occupies a thread block, and within the thread block, work is parallelized as indicated in the above diagram. }
  \label{fig:gpu}
\end{figure}

\newcommand{\ParScan}[3]{
  \State $ #1 \gets \mathbf{prefix\mbox{-}scan}_{#3}\!\left( #2 \right) $%
}
The feasibility of faster than real-time sampling-based NMPC depends on rapid evaluation of multiple instances of the underlying unsteady fluid-dynamics model. To this end, we implement a particle-wise GPU parallel implementation of the VPM.  Parallelization over both particles and models allows us maximum throughput and saturation of GPU streaming multiprocessors.
Each particle-particle interaction is computed in a CUDA thread, and aerodynamic loads are resolved simultaneously across model instances. Each model instance is allocated a $256$ thread block. The linear system in Eq. \ref{eq:system} is solved with the LU factorization routine provided by cuBLAS \cite{cuda12}.  The parallelization scheme is depicted in Fig. \ref{fig:gpu}. We leverage a parallel prefix scan for the summations in Eq. \ref{eq:wake-contrib} and Eq. \ref{eq:convect}.

Because runtime performance of parallel dynamics rollouts is of paramount importance to enable real-time replanning, we evaluate runtime on a RTX 5080 GPU by stepping the dynamics 80 times (the approximate length of our perch maneuver at $\Delta t = 0.01$), allowing for a maximum of 60 particles before applying aggregation. The evaluation time by batch size is depicted in Table \ref{tab:batch_runtime}. These times are sufficient for 10 Hz replanning on a modern GPU, depending on the desired number of MPPI iterations per replanning cycle.
\begin{table}[b]
\centering
\begin{tabular}{|c|c|}
\hline
\textbf{Batch Size} & \textbf{Runtime (ms)} \\ \hline
1   & 8.0  \\ \hline
128 & 10.3 \\ \hline
256 & 12.2 \\ \hline
512 & 16.3 \\ \hline
1024 & 27.2 \\ \hline
\end{tabular}
\caption{Runtime vs. batch size for VPM rollouts.}
\label{tab:batch_runtime}
\end{table}
\section{Real-Time Planning and Control}
To achieve precision post-stall perching with the fixed-wing glider, we combine the unsteady aerodynamics model developed in Sec. \ref{sec:unsteady} with a stochastic NMPC framework. Our NMPC framework consists of a trajectory optimizer and trajectory tracking controller which operate directly on samples from the dynamics model. Our control approach has two distinct advantages: 1) it can reason predictively in real-time about the unsteady aerodynamics over the entire maneuver to improve perching performance and 2) it does not need to compute gradients to execute the local closed-loop policy.
\subsection{Trajectory Optimization}
Our high angle-of-attack, post-stall, perching landing trajectory is obtained through model predictive path integral control (MPPI) \cite{mppi} minimizing the cost function 
\begin{align} \label{eq:mppi-cost}
    J = (\vect{x}_{g, f} - \vect{x}_d)^T \vect{Q} (\vect{x}_{g,f}- \vect{x}_d)
\end{align}
where $\bx_{g,f}$ is the final glider trajectory state, and $\vect{x}_d = \begin{bmatrix} 3.5, 0, \frac{\pi}{4}, 0, \frac{1}{2}, -\frac{1}{2}, 0\end{bmatrix}^T$ is the desired perch state.
MPPI is a stochastic shooting method that samples control input sequences, reweights them according to the resultant costs, and iterates until convergence. Let a control trajectory of length $N$ be given as $\boldsymbol{\xi} =\begin{bmatrix} \bu_0^T&\bu_1^T&\dots \bu_N^T \end{bmatrix}^T$. Control trajectories are sampled from normal distributions $\boldsymbol{\xi}_i \sim \mathcal{N}(\boldsymbol{\mu}, \vect{\Sigma})$ and re-weighted according to an exponentially weighted average of trajectory costs $J_i$ with temperature parameter $\lambda$ and normalization factor $\eta$. The optimal control sequence is given as
\begin{align}
    \boldsymbol{\hat{\xi}^*} = \eta \sum_{i=1}^Be^{-\frac{J_i-J_\text{min}}{\lambda}} \vect{\bxi}_i.
\end{align}
The initial glider state is $\bx_{g,0} = \begin{bmatrix} 0 & 0 & 0 & 0 & 7 & 0 & 0\end{bmatrix}^T$.

\subsection{Trajectory Tracking}
Numerical integration of the VPM yields noisy forces and consequently, non-smooth system trajectories. This limits the applicability of gradient-based trajectory tracking controller synthesis methods (such as time-varying LQR) that naively rely on finite-differencing through the VPM. Thus, in order to design a local feedback policy, we employ a least-squares technique, similar to \cite{gupta2026stochastic}, to synthesize a local time-varying linear system from rollouts. Additionally, this allows us to select only the directly observable subset of the high dimensional state space for the purpose of feedback.

Consider a nominal trajectory consisting of a sequence of states $\bx_k$ and inputs $\bu_k$ for times $k=0,1\ldots N$. We generate a set of $K$ nearby, perturbed initial states  $\mathcal{X}^0 = \{\vect{x}^i\}_{i=1}^K \sim \mathcal{N}(\boldsymbol{\bx}_0, \boldsymbol{\Sigma}_x)$ 
and nearby, perturbed control input trajectories  $\mathcal{U}^k = \{\vect{u}^i\}_{i=1}^K \sim \mathcal{N}(\boldsymbol{\bu}_k, \boldsymbol{\Sigma}_u)$. We use superscript $i$ to denote the set corresponding to the $i^{th}$ time. Sets $\mathcal{X}^2, \mathcal{X}^3, \dots, \mathcal{X}^N$ are generated via dynamics rollouts from $\mathcal{X}^0$. Using a first order Taylor series approximation, at time $k$ we can write 
\begin{align}
    \dot{\vect{x}}^i \approx \vect{f}(\bx_k, \bu_k) 
    + \frac{\partial \vect{f}}{\partial \vect{x}} \, (\vect{x}^i - \bx_k) 
    + \frac{\partial \vect{f}}{\partial \vect{u}} \, (\vect{u}^i - \bu_k)
\end{align}
for $i=0,1\ldots K$.

Letting $\Delta \vect{x}^i = \vect{x}^i - \bx_k$ with $\vect{x}^i \in \mathcal{X}^k$  
and $\Delta \vect{u}^i = \vect{u}^i - \bu_k$ with $\vect{u}^i \in \mathcal{U}^k$ it follows that
\begin{align*}
    &\begin{bmatrix}
    \dot{\vect{x}}^1 - \dot{\bx}_k & \dots & \dot{\vect{x}}^K - \dot{\bx}_k
    \end{bmatrix} \\
    &= \frac{\partial \vect{f}}{\partial \vect{x}} 
    \begin{bmatrix}
    \Delta \vect{x}^1 & \dots & \Delta \vect{x}^K
    \end{bmatrix}  +
    \frac{\partial \vect{f}}{\partial \vect{u}} 
    \begin{bmatrix}
    \Delta \vect{u}^1 & \dots & \Delta \vect{u}^K
    \end{bmatrix} 
    \\
    &= \begin{bmatrix} 
    \frac{\partial \vect{f}}{\partial \vect{x}} &
    \frac{\partial \vect{f}}{\partial \vect{u}}
    \end{bmatrix}
    \begin{bmatrix}
        \Delta \vect{x}^1 & \dots & \Delta \vect{x}^K \\
        \Delta \vect{u}^1 & \dots & \Delta \vect{u}^K
    \end{bmatrix}
\end{align*}
From the above we can now construct the following least-squares problem to determine the system Jacobians:
\begin{align} \label{eq:synthesis}
    \arg\min_{ 
    \frac{\partial \vect{f}}{\partial \vect{x}},
    \frac{\partial \vect{f}}{\partial \vect{u}} }
    \left\|
    \begin{bmatrix}
    (\dot{\vect{x}}^1 - \dot{\bx}_k)^T \\ \dots \\ (\dot{\vect{x}}^K - \dot{\bx}_k)^T
    \end{bmatrix}  - 
    \begin{bmatrix}
        \Delta \vect{x}^{1T} & \Delta \vect{u}^{1T} \\
        \dots & \dots \\
        \Delta \vect{x}^{KT} & \Delta \vect{u}^{KT}
    \end{bmatrix}
    \begin{bmatrix} 
    \frac{\partial \vect{f}}{\partial \vect{x}}^T \\
    \frac{\partial \vect{f}}{\partial \vect{u}}^T 
    \end{bmatrix}
    \right\|_2
\end{align}

This procedure is applied to determine the system matrices for the 7 directly observable glider states $\bx_g = \begin{bmatrix} x & z & \theta & \phi & \dot{x} & \dot{z} & \dot{\theta}\end{bmatrix}^T$ and the elevator rate $\bu= \dot{\phi}$. Crucially, we exploit the known sparsity structure of the glider Jacobians to solve only for the unknown $3 \times 5$ block of matrix $\vect{A}_k=\frac{\partial \vect{f}}{\partial \vect{x_g}}$, $\vect{\hat{A}}$, and $3\times 1$ block of matrix $\vect{B}_k=\frac{\partial \vect{f}}{\partial \vect{u}}$, $\hat{\vect{B}}$, via Eq~\ref{eq:synthesis}, as shown below:
\begin{align}
\vect{A}_k =  \begin{bmatrix}
\vect{I}_{\text{3x3}} & \vect{0}_\text{3x4} \\
\vect{0}_\text{1x3} & \vect{0}_\text{1x4} \\
\vect{0}_\text{3x2} & \hat{\vect{A}}
\end{bmatrix},
\quad
\vect{B}_k = \begin{bmatrix}
\vect{0}_\text{3x1}\\
1 \\
\hat{\vect{B}}
\end{bmatrix}.
\end{align}
This prevents the solution from containing spurious correlations between states. 

Once the system matrices have been identified, we compute feedback gains via the time-varying linear quadratic regulator (TVLQR) by numerically integrating the TVLQR Riccati equations
\newcommand{\matr}[1]{\mathbf{#1}} 
\begin{align}
& \matr{S}_{k-1} = \matr{A}_k^T\matr{S}_k\matr{A}_k-...
\nonumber\\&\matr{A}_k^T\matr{S}_k(\matr{R}_c+\matr{B}_k^T\matr{S}_k\matr{B}_k)^{-1}\matr{B}_k^T\matr{S}_k\matr{A}_k+\matr{Q_c}
\end{align}
backwards in time from $\vect{S}_N=\vect{Q}_f$, where $\bS_k$ is the cost-to-go matrix. Time-dependent feedback gains are given by
\begin{align} \label{eq:tvlqr-gain}
 &\matr{H}_k = (\matr{R}_c + \matr{B}_k^T\matr{S}_{k+1}\matr{B}_k)^{-1}(\matr{B}_k^T\matr{S}_{k+1}\matr{A}_k). 
\end{align}The control input is calculated by 
\begin{align} 
& \vect{u}_k = -\matr{H}_k(\vect{x}_{g,k}-\vect{x}^d_k)+\vect{u}^d_k.
\end{align} 

\subsubsection{Feedback Control Performance}
\begin{figure}[tbh]%
  \centering
    \includegraphics[clip, width=1.0\columnwidth]{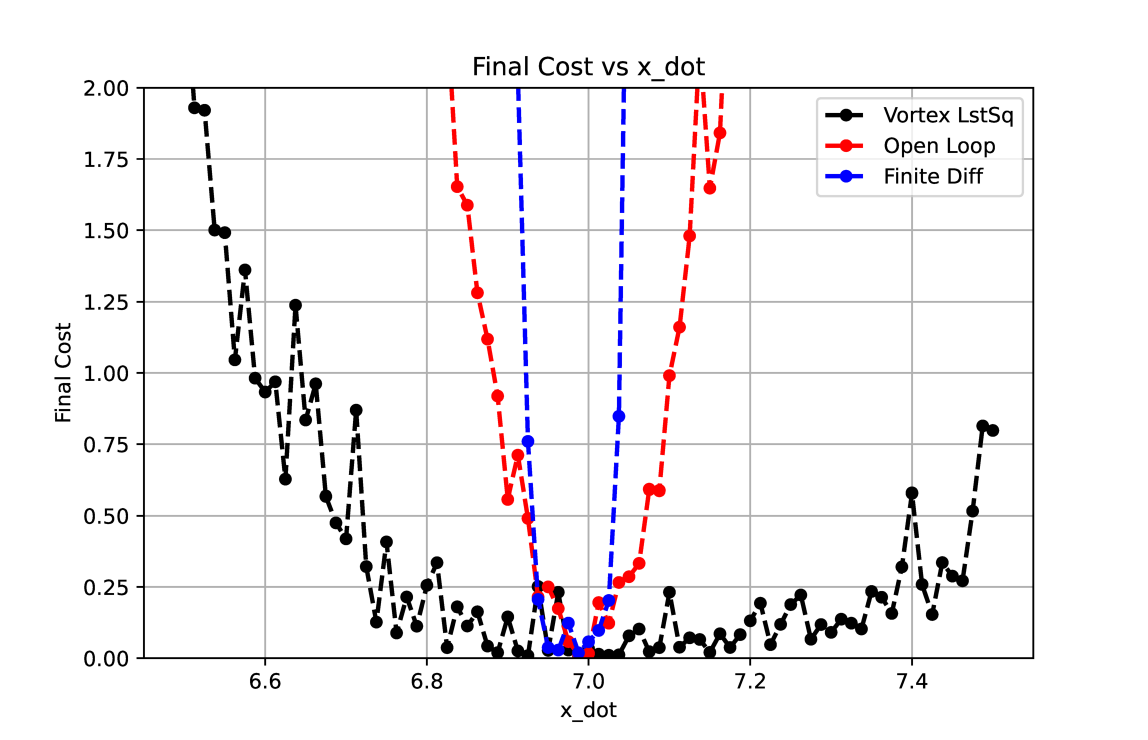}
    \caption{Initial condition sweep vs. final cost for $\dot{x}$. 7 m/s is the nominal initial condition. The least squares policy has the largest region of attraction, while the policy constructed from finite differencing shows a region of attraction even smaller than the open-loop policy.}
  \label{fig:ic-sweep}
\end{figure}

We validate the efficacy of the least squares system synthesis and TVLQR policy by using numerical simulations to evaluate the final cost of the trajectory tracking policy under initial condition perturbations. An example initial condition sweep over horizontal velocity is shown in Fig.~\ref{fig:ic-sweep}. In this scenario, the feedback policy significantly outperforms the open loop result, giving us a region of attraction of $\pm$ 0.4 m/s for a 5 cm final perch error. Similar sweeps were run across other state slices to determine the best parameters for the noise covariances and TVLQR gains. We also observe that the finite difference policy performs worse than the open loop system due to the aforementioned noise. The policy construction, including the least square system synthesis and solving the TVLQR Riccati equation takes 9 milliseconds.

\subsection{Nonlinear Model Predictive Control}
\begin{figure}[tbh]%
  \centering
    \includegraphics[clip, width=1.0\columnwidth]{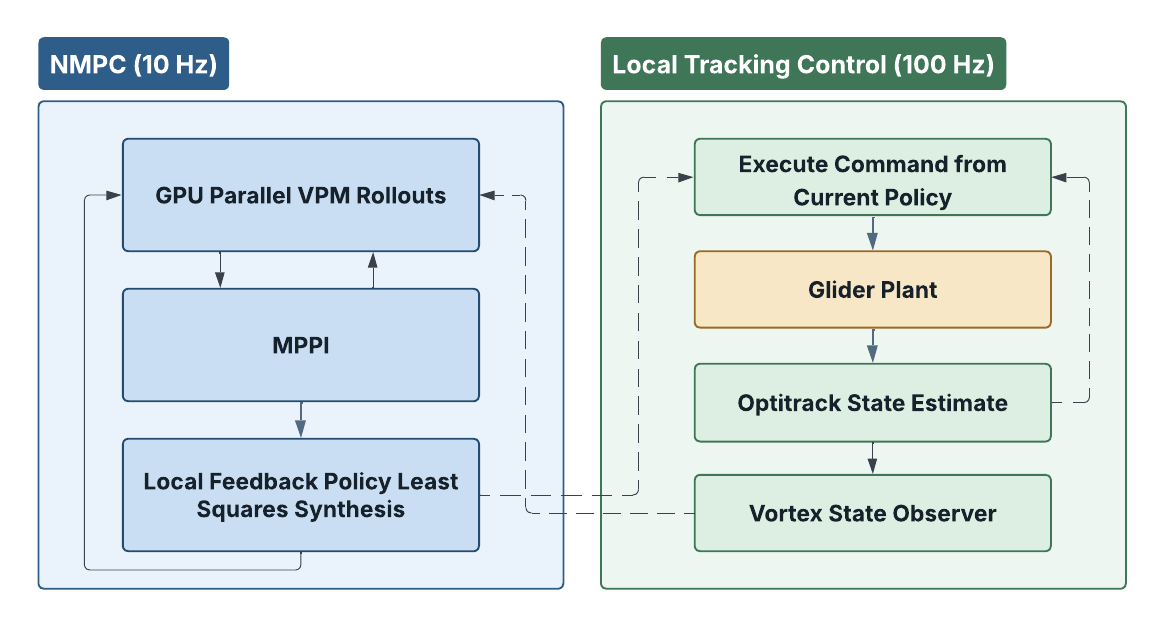}
    \caption{Overall controller architecture. Solid arrows indicate control flow, and dashed arrows indicate information passing.  }
  \label{fig:flowchart}
\end{figure}

We combine our sampling-based approaches for trajectory optimization and local policy synthesis together to create a real-time feedback NMPC algorithm \cite{grandia2019feedback}. Given a vortex state, policy distribution, and replanning time, the algorithm proceeds as follows:
\begin{enumerate}
\item Project the current vortex state forward in time using the current closed loop policy to compensate for the replanning time.
\item Rollout parallelized dynamics model with previous policy distribution and apply MPPI reweighting.
\item Optimize the time-varying linear system and construct a new feedback policy using TVLQR. 
\end{enumerate}
Because we require the current wake state of the glider, our algorithm maintains a vortex model instance that is simulated forward with the observed rigid body state at every controller step. Our NMPC approach is implemented using two parallel threads: a control thread which executes the feedback policy and a planning thread which generates the nominal trajectory and local feedback policy. This approach is summarized in Fig.~\ref{fig:flowchart}.

\algblockdefx{ParFor}{EndParFor}[1]{\textbf{parallel for} #1}{\textbf{end parallel for}}

\section{Perching Maneuver Experiments} 
To explore the performance of our NMPC approach, we consider a modified version of the fixed-wing perching problem outlined in \cite{moore2014robust}, the ``disturbed'' fixed-wing perching problem. In this modified scenario, a translating vortex ring disturbance is added in the vicinity of the perch location (see Fig. \ref{fig:setup}). Not only is the unsteady aerodynamics model useful for capturing the vortex shedding associated with the high angle-of-attack perching maneuver, but it can also account for the inherently unsteady vortex ring disturbance. 

To evaluate the performance of the NMPC algorithm on the perching maneuver, we conduct the following experiments: 
\begin{itemize}
    \item Post-stall perching without a disturbance.
    \item ``Uncompensated'' post-stall perching with a disturbance, where the vortex ring is physically present, but the planner is without knowledge of the disturbance.
    \item ``Compensated'' post-stall perching with a disturbance, where an estimate of the vortex ring state is included during trajectory planning.
\end{itemize}
To model the vortex ring disturbance using our VPM, two additional vortices of equal magnitude but opposite sign can be injected into the wake and convected with the rest of the particles. In order to prevent non-physical interaction of the disturbance vortices with the bound airfoil vortices, we terminate the disturbance at the wing. We achieve this by a line-to-line intersection check: The chord line is inflated by 2 percent in the direction opposite of the normal, which is checked against the line formed by the two ring vortices.

\subsection{Experimental Set-up} \label{sec:exp-setup}
The controller is run on a PC with an RTX 5080 graphics card and a Ryzen 9 9950X CPU. The vehicle state is determined using a motion capture system streaming at 120~Hz. The NMPC replanning frequency is set to 10~Hz, and all vortex models are stepped at $\Delta t = 0.01 \text{s}$. The nominal trajectory has 77 steps, and we use an MPPI batch size of 256 with 3 iterations per replanning cycle.  We use $\bQ = diag(\begin{bmatrix}10, 10, 1, 0, 0.2, 0.2, 0.2\end{bmatrix})$ for the NMPC, and $\bQ_{f,c} = diag(\begin{bmatrix}400, 400, 10, 1, 1, 1, 1\end{bmatrix})$, $\bQ_c = diag([0.1, 0.1, 5.0, 0.1, 0.1, 0.1, 5.0])$, and ${\bf R}_c = 0.01$ for TVLQR. Commands are wirelessly sent to the aircraft via a Futaba T12K RC transmitter. The PC sends serial commands to an Arduino Nano that relays them using Pulse Position Modulation (PPM) to the transmitter's trainer port.

\begin{figure}[tbh]%
  \centering
    \includegraphics[clip, width=1.0\columnwidth]{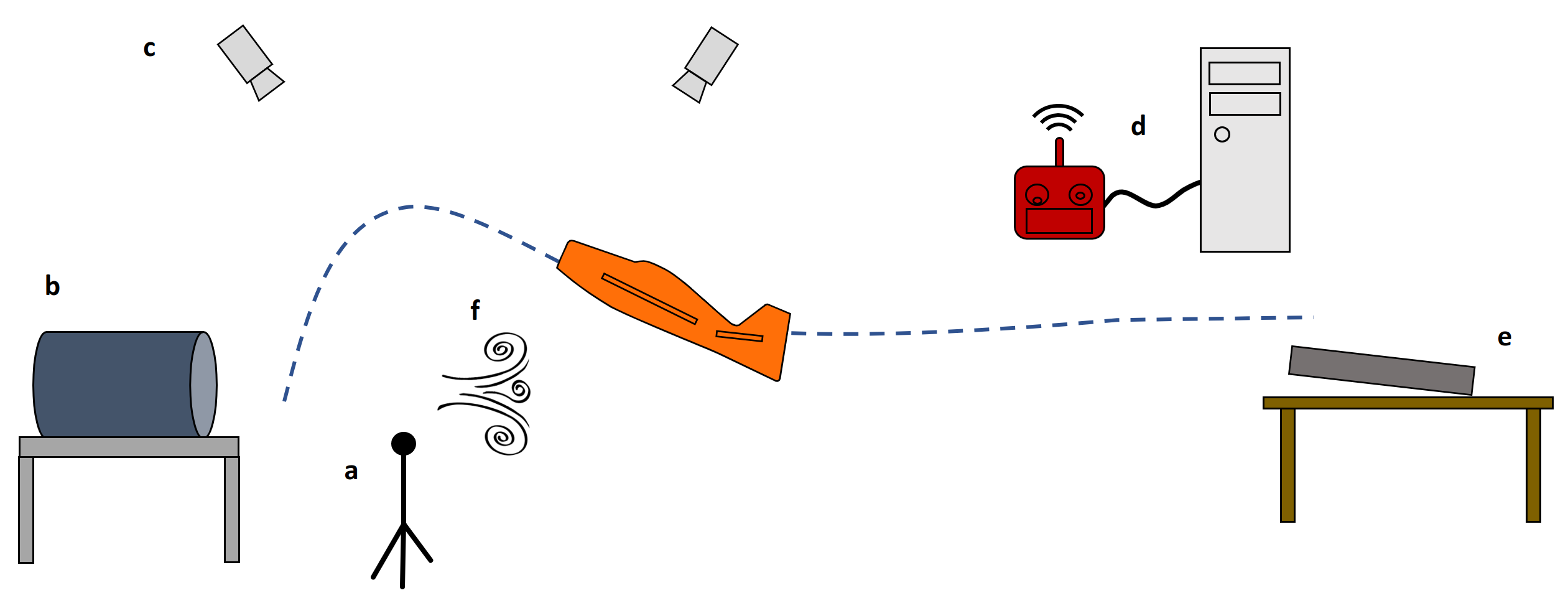}
    \caption{The experimental setup: a) ground pressure sensor b) vortex cannon. c) motion capture cameras. d) PC and transmitter. e) launcher system. f) vortex ring disturbance.}
  \label{fig:setup}
\end{figure}

The disturbance is generated by a vortex ring cannon built from a 32-gallon plastic trashcan with a 28-cm diameter cutout in the bottom and a tarp attached by bungee cords. The cannon and glider are launched by human operators. In the scope of this work, we are primarily concerned with evaluating controller performance given additional vortex information, as opposed to vortex state estimation. Consequently, we characterize the vortex ring parameters offline using two statically mounted pressure sensors spaced approximately 2~m apart. The vortex ring is released, and the time of impact is measured at each sensor to obtain a velocity estimate of 7.5~m/s for the gust. It should be noted that this is an extremely significant disturbance in the context of our setup since the glider velocity along the trajectory is always less than 7~m/s. We assume the vortex cores are separated by the diameter of the cutout in the cannon. The vorticity can be computed using Eq. (\ref{eq:regularized_kernel}) for the regularized kernel. During experiments, only one offboard pressure sensor is used, and the vorticity, separation, and height relative to the sensor are held constant at the values recorded from the offline measurement. When the pressure sensor detects a 15 pascal increase from the baseline reading, the position of the vortex is injected into the vortex model. The full setup is depicted in Fig.~\ref{fig:setup}.
\subsection{Experimental Results} \label{sec:hw-result}
\begin{figure}[tbh]%
  \centering
    \includegraphics[clip, width=1.0\columnwidth]{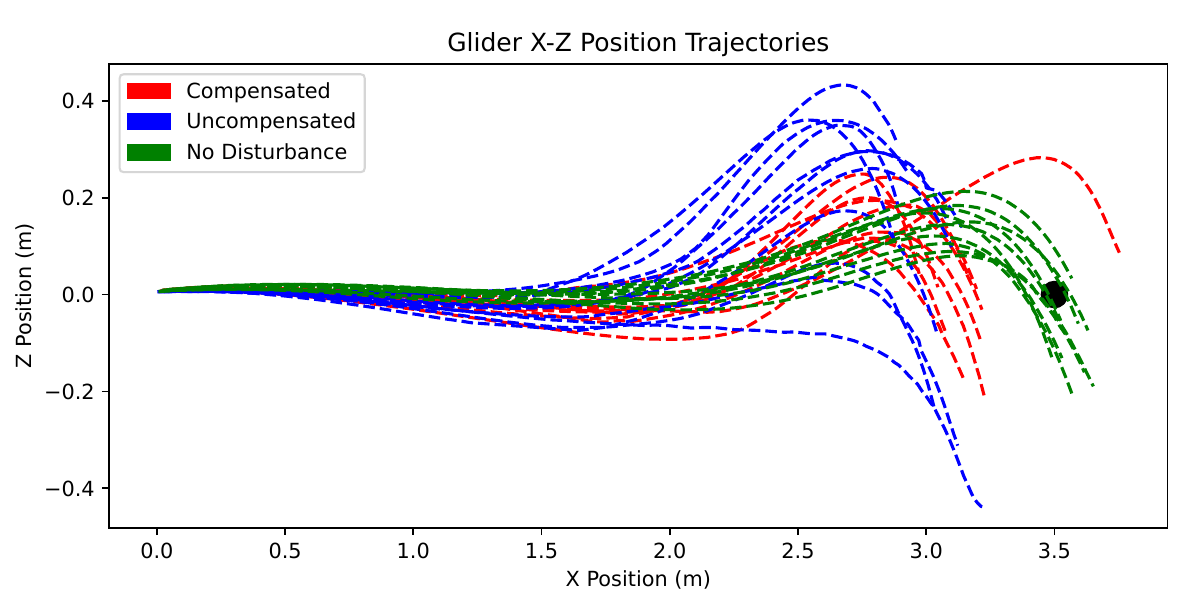}
    \caption{Position trajectories for all trials. The trials without disturbance (green) show best perching performance, but the disturbed trials with compensation (red) show better final position error and trajectory distribution variance than without (blue).}
  \label{fig:pos-series}
\end{figure}

\begin{figure}[tbh]%
  \centering
    \includegraphics[clip, trim={0 40 0 0}, width=0.9\columnwidth]{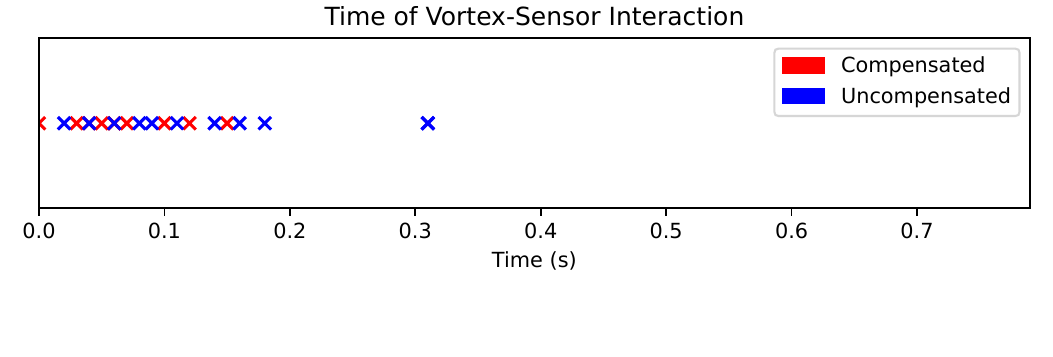}
    \caption{Pressure sensor-vortex interaction time. With the exception of one outlier, the vortex-sensor interactions are comparable across compensated and uncompensated trials.}
  \label{fig:interaction}
\end{figure}

\begin{figure}[tbh]%
  \centering
    \includegraphics[clip, width=0.9\columnwidth]{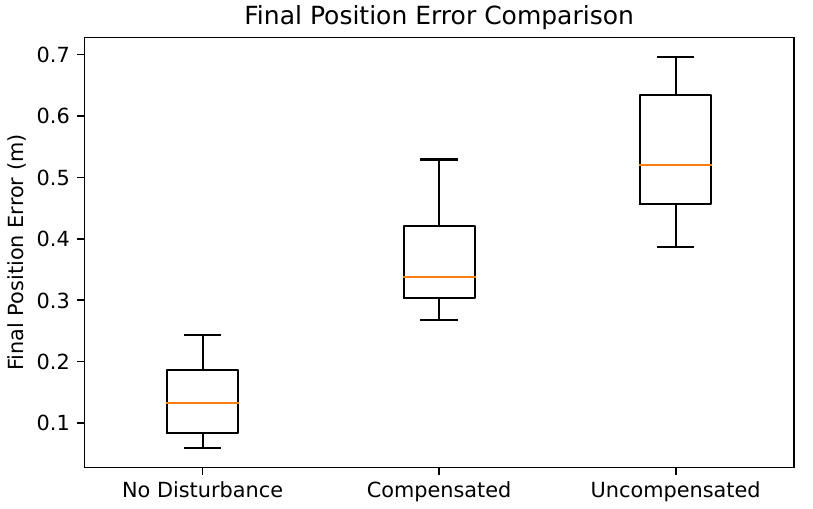}
    \caption{Final position error for 10 trials of each experiment series. Vortex disturbance compensation outperforms uncompensated trials in final position error.}
  \label{fig:nmpc}
\end{figure}

\begin{figure}[tbh]%
  \centering
    \includegraphics[clip, width=1.0\columnwidth]{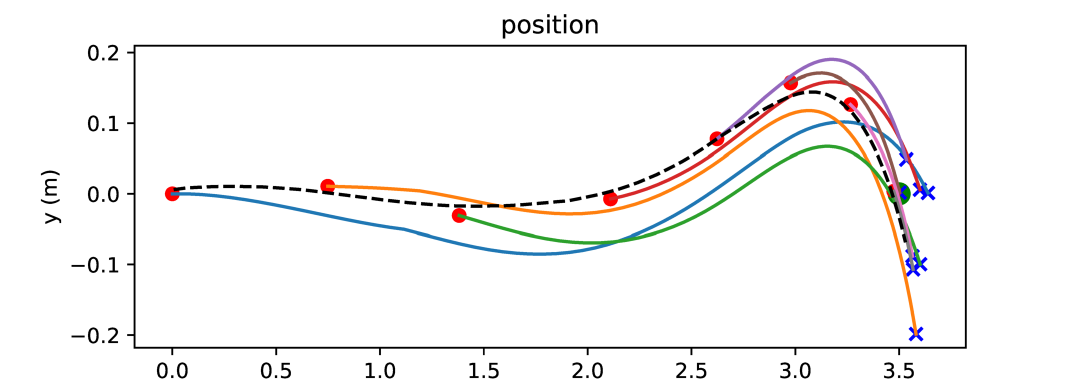}
    \caption{Example of a single hardware perching trial using NMPC. The black dashed line is the true position, red dots show projected positions, and colored lines are replan trajectories.}
  \label{fig:replan}
\end{figure}

\begin{figure}[tb]%
  \centering
    \includegraphics[clip, trim={0 0 0 0}, width=0.9\columnwidth]{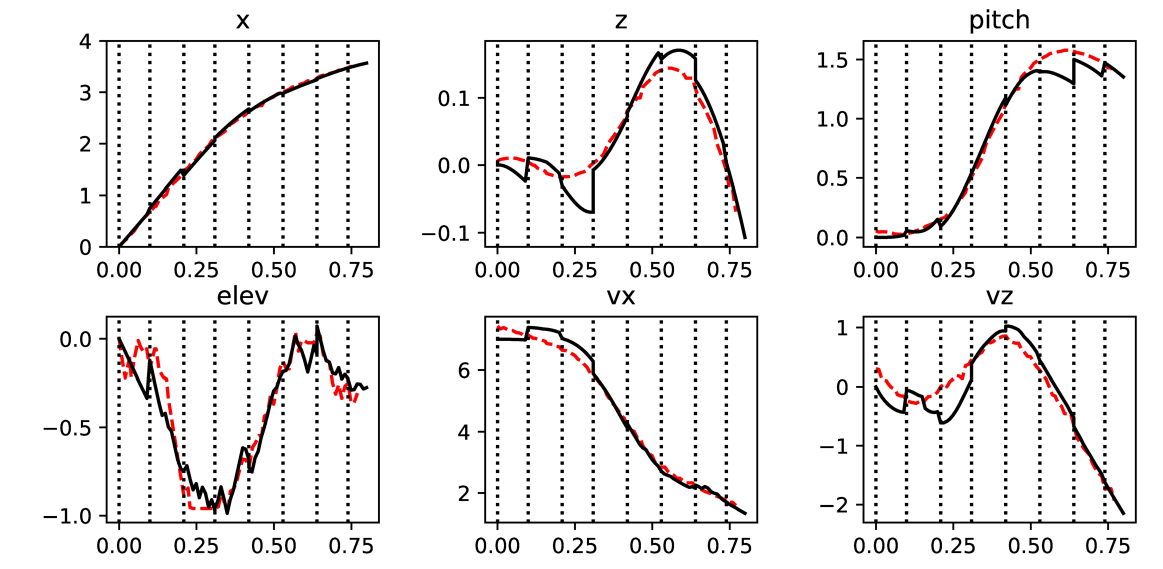}
    \caption{Example of state tracking with TVLQR. Red dashed lines are true state trajectories, and black line shows replanned reference. Vertical dotted lines show replan intervals.}
  \label{fig:state-err}
\end{figure}

We ran 10 perching trials for each of the three experiments detailed above. Representative NMPC perching results can be seen in Fig.~\ref{fig:replan} and Fig.~\ref{fig:state-err}. In Fig.~\ref{fig:nmpc} we see that NMPC without a disturbance results in the lowest terminal position error, but NMPC with disturbance compensation demonstrates lower error than that without. This indicates that modeling the disturbance in the fluid state improves controller performance. The position trajectories for all trials colored by experiment are shown in Fig.~\ref{fig:pos-series}.  The vortex cannon is operated by a human operator, which leads to some variation in the impact velocity and forces experienced by the glider. Fig.~\ref{fig:interaction} shows the vortex-pressure sensor interaction times, which indicate that, with the exception of one outlier, the timing of the manual vortex release was comparable across compensated and uncompensated trials.


\subsection{Model Validation}
\begin{figure}[tbh]%
 \centering
   \includegraphics[clip, width=0.9\columnwidth]{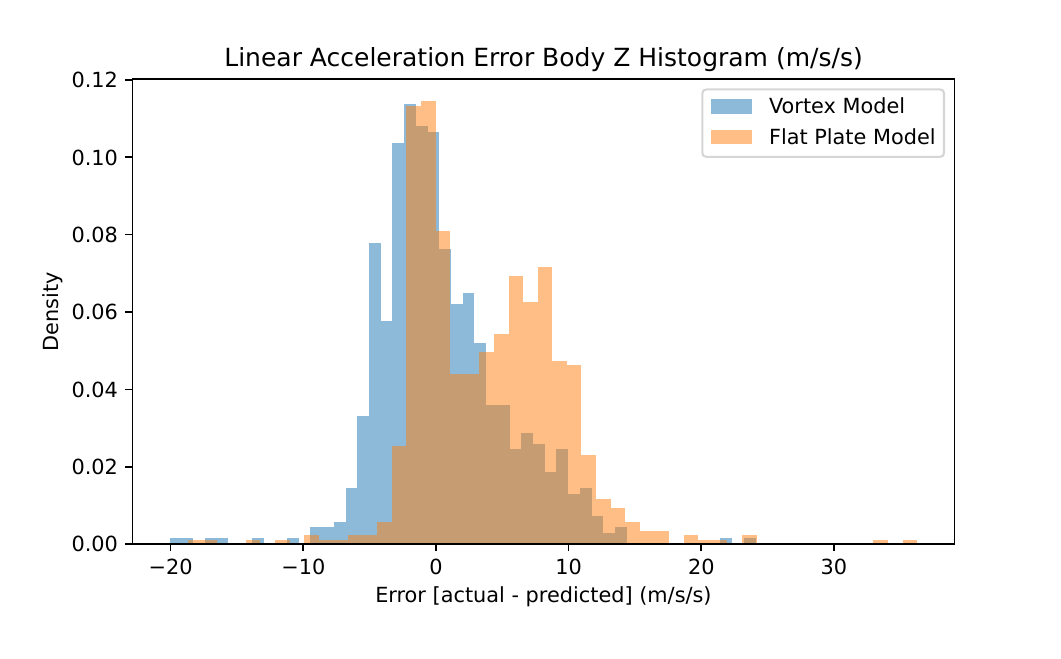}
   \caption{Body-z acceleration errors for VPM and quasi-static flat plate model. The vortex model provides acceleration errors with a lower mean and variance.}
  \label{fig:zerr}
\end{figure}

We validate the predictive accuracy of the VPM by computing the body-Z acceleration errors against the measured values from motion capture for the undisturbed perching trials. The error histogram is seen in Fig. \ref{fig:zerr}. We also include the quasi-static flat plate theory force predictions for reference. The VPM errors are more axis-symmetric and exhibit less spread as shown in table \ref{tab:error}.

\begin{table}[tbh]
\centering
\begin{tabular}{|c|c|c|}
\hline
\textbf{Model} & \textbf{Mean $\mathbf{m\,s^{-2}}$} & \textbf{Std. Dev. $\mathbf{m\,s^{-2}}$} \\ \hline
VPM   &   0.5 & 4.7\\ \hline
Flat Plate & 3.9 & 5.4 \\ \hline
\end{tabular}
\caption{Mean and standard deviation for body-z accelerations for VPM and flat plate model.}
\label{tab:error}
\end{table}

\section{DISCUSSION AND CONCLUSION}
In this work, we proposed a GPU parallelized VPM and associated stochastic optimal control strategies for real-time planning of fixed-wing maneuvers in unsteady flow conditions. Using solely trajectory samples from the lightweight fluid-dynamics model, we achieve online motion planning and feedback policy construction. Our hardware results demonstrate that by reasoning over the fluid state, it is possible to improve maneuver execution performance in a heavily-disturbed, time-varying flow. We intend to carry out further investigations for online estimation of the wake state and disturbance parameters using onboard pressure sensors.







\FloatBarrier

\bibliographystyle{IEEEtran}
\bibliography{references}

\end{document}